\documentclass[conference]{IEEEtran} 
\usepackage{cite}

\ifCLASSINFOpdf
\else
\fi
\ifCLASSOPTIONcompsoc
  \usepackage[caption=false,font=normalsize,labelfont=sf,textfont=sf]{subfig}
\else
  \usepackage[caption=false,font=footnotesize]{subfig}
\fi
\usepackage{url}

\usepackage{float}
\usepackage{amssymb}
\usepackage{amsmath}
\usepackage{amsthm}
\usepackage{tikz}
\usepackage{graphicx}
\usepackage{pgfplots}
\usepackage{pstool}  
\usepackage[linesnumbered,algoruled,boxed,lined]{algorithm2e}
\usepackage{algcompatible}
\usepackage{threeparttable}
\usepackage[strict]{changepage}
\usepackage{bbm}
\usepackage{bm}
\usepackage{comment}
\usepackage{hhline}

\begin{document}
%
\title{Machine Learning in Downlink Coordinated Multipoint in Heterogeneous Networks}


\author{\IEEEauthorblockN{Faris B.~Mismar and Brian L.~Evans}

\IEEEauthorblockA{Wireless Networking and Communications Group, The University of Texas at Austin, Austin, TX 78712 USA}

}

\maketitle

\begin{abstract}
We propose a method for downlink coordinated multipoint (DL CoMP) in heterogeneous fifth generation New Radio (NR) networks.  The primary contribution of our paper is an algorithm to enhance the trigger of DL CoMP using online machine learning.  We use support vector machine (SVM) classifiers to enhance the user downlink throughput in a realistic frequency division duplex network environment.  Our simulation results show improvement in both the macro and pico base station downlink throughputs due to the informed triggering of the multiple radio streams as learned by the SVM classifier.
\end{abstract}

%
\IEEEpeerreviewmaketitle

\section{Introduction}
The demand for data traffic over cellular networks continues to increase with emphasis on low latency and high reliability. Heterogeneous networks are an important solution to the problem of increase in capacity demand.  In heterogeneous networks, pico base stations are deployed with the existing macro base stations.  
The downlink \textit{coordinated multi-point} (DL CoMP) operation was first introduced in \mbox{3gpp Rel 11} for \textit{long term evolution advanced} (LTE-A) networks.  It was a feature that improved data rates coverages and cellular capacity at cell edge using a fiber backhaul \cite{3gpp36819, 7947086}.   
DL CoMP was further enhanced in 3gpp Rel~13 (eCoMP) with fast \textit{channel state information} (CSI) acquisition messages being sent between the base stations involved.  DL CoMP will play an important role in the \textit{fifth generation of wireless communications} (5G) air interface which is also known as \textit{New Radio} (NR) \cite{6815892}.

CoMP was studied extensively in several papers \cite{huq,7432047,7938355} with solutions offered through convex optimization, Markov chain based models, and queuing theory.  Different from these papers, we employ machine learning on the joint transmission scheme where the UE is likely to receive data from multiple streams.

In this paper, we focus on the \textit{joint processing} scheme of CoMP in the downlink direction, where the receiver is the \textit{user equipment} (UE).  Spatially multiplexed data streams transmitted by the \textit{base station} (BS) are available at more than one transmission point.  These points (or base stations) form the CoMP \textit{cooperating set}.  This effectively forms a distributed \textit{multiple input multiple output} (MIMO) channel with streams from each BS in the CoMP cooperating set. 

\begin{figure}[!t]
\centering
\includegraphics[scale=0.76]{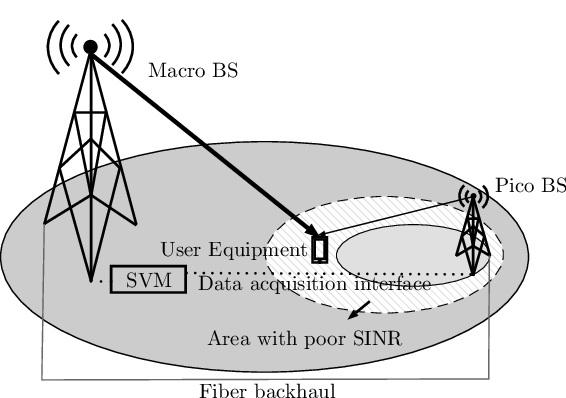}
\caption{Joint processing and support vector machine (SVM) in a coordinated multipoint heterogeneous network.}
\label{fig:dlcomp}
\end{figure}



Our objective is to improve the CoMP joint processing distributed MIMO performance.  To achieve this objective, we propose an online supervised machine learning based algorithm which acquires physical layer data from the connected UEs within the channel coherence time in a radio frame.  This algorithm can reside in a centralized location as part of a \textit{self-organizing network} (SON) or in an edge compute node at the BS. We use a minimalistic set of learning features to keep the time and space complexity in polynomial order.  The overall view is in Fig.~\ref{fig:dlcomp}.

Our main contributions are as follows:
\begin{enumerate}
\item Demonstrate that a machine learning model can improve the performance of joint processing CoMP triggering in a realistic environment.
\item Increase the user throughput in a heterogeneous network as a result of learning improved triggering conditions of CoMP compared to the industry-compliant baseline.
\end{enumerate}

\section{System Model} \label{sec:system_model}

The system is composed of two modules:
\begin{itemize}
\item An inter-site CoMP operation in a heterogeneous network composed of macro and pico base stations connected with optical fiber. 
\item A machine learning algorithm using a \textit{support vector machine} (SVM) classifier to derive improved triggering point for CoMP to operate if applicable.
\end{itemize}

\subsection{Radio Environment}\label{sec:radio_environment}
Our setup for the macro base stations uses hexagonal cellular geometry.  We use pico base stations for densification of the macro coverage in an urban environment.  Non-stationary UEs with multiple antennas are randomly placed and uniformly distributed in the service area.  The base stations are the transmitters and the UEs are the receivers.  We use 5G NR as a multi-access wireless network in the sub-6 GHz frequency range and the \textit{frequency division duplex} (FDD) mode of operation (i.e., no channel reciprocity).  We can therefore write the signal of an arbitrary UE $q$ as
\begin{equation}
\mathbf{r} = \mathbf{H}\mathbf{s} + \mathbf{v}, \qquad q= 1, \ldots, Q.
\label{eq:channel}
\end{equation}
The subscript $q$ is dropped for ease of notation.  Here, $\mathbf{r}  \in \mathbb{C}^{n_r}$ is the received signal (i.e., at the UE side). $\mathbf{H}\in \mathbb{C}^{n_r\times n_t}$ is the Rayleigh fading channel for the $q$-th UE with independent identically distributed (i.i.d.) circularly symmetric standard complex Gaussian entries.  Further, $\mathbf{s}  \in \mathbb{R}^{n_t}$ is the transmitted signal, and $\mathbf{v} \in \mathbb{C}^{n_r}$ is the noise plus interference both of which are also assumed circularly symmetric Gaussian, a baseline practice even in 5G systems \cite{8375976}.  Finally, $n_r$ and $n_t$ are the number of receive and transmit streams respectively such that the maximum number of streams $n_s^\textrm{max}\triangleq\min(n_r,n_t)$.

Since 5G NR is based on \textit{orthogonal frequency division multiplexing} (OFDM), we choose \textit{zero-forcing} (ZF) channel equalization.  This sets the inter-cellular interference to zero and allows us to deal with Gaussian noise. Hence, SINR and SNR can be used interchangeably. We write our ZF equalizer $\mathbf{W}_\text{ZF}\in\mathbb{C}^{n_t\times n_r}$ for the $q$-th UE as
\begin{equation}
\mathbf{W}_\text{ZF} = (\mathbf{H}^\text{H} \mathbf{H})^{-1}\mathbf{H}^\text{H}
\end{equation}
with $(\cdot)^\text{H}$ denoting the Hermitian transpose operation.  This allows the UE to obtain an estimate of the transmitted signal through pre-multiplication of $\mathbf{W}_\text{ZF}$ into \eqref{eq:channel}.  The parameters of the radio environment are listed in Table~\ref{table:parameters}.

Now, we can compute the SNR per receive stream, observe respective transmission block errors, and find out the bitrates using the simulator \cite{VLS-2016}.  We expect that as the number of receive streams increases, the block error rate increases, but the bitrates per stream also increase.  The net change is an increase in the user throughput distribution.

\subsection{Machine Learning} \label{sec:machinelearning}

\begin{figure}[!t]
\begin{adjustwidth}{-.1in}{0cm}
\centering
\resizebox{.7\linewidth}{!}{\input{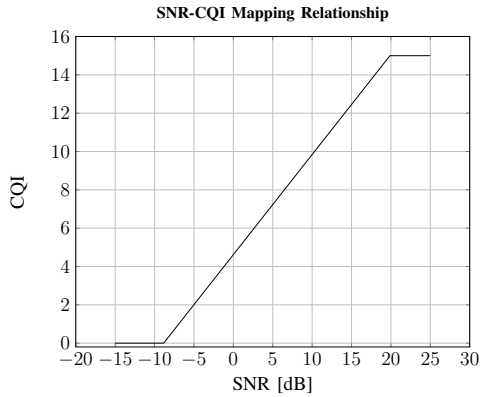}} 
\vspace*{-.1in}
\end{adjustwidth}
\caption{Used relationship between signal to noise ratio (SNR) and call quality indicator (CQI) \cite{VLS-2016, mehlf}.}
\label{fig:snrcqi}
\end{figure}

We use the SVM classifier \cite{Cortes95support-vectornetworks} in the implementation of this algorithm.  We define the learning features in a matrix $\mathbf{X}$ as listed in Table~\ref{table:mlfeatures}.  These features are collected from all the UEs in the CoMP cooperating set during the time duration of $T_\text{CoMP}$.  This duration cannot exceed either the channel coherence time or the radio frame duration measured in \textit{transmit time intervals} (TTIs).   From the CSI in NR \cite{3gpp38215}, we choose a linearly mapped version of the signal to noise and interference ratio (CSI-SINR) and CSI \textit{reference symbol received power} (CSI-RSRP) as $\mathbf{x_1}$ and $\mathbf{x_2}$ respectively.   This linearly mapped version of the NR CSI-SINR resembles what LTE/LTE-A calls the \textit{channel quality indicator} (CQI) \cite{3gpp36213} and shall be the name we use here, as shown in Fig.~\ref{fig:snrcqi}. 

We observe the transport \textit{block error rate} (BLER), computed at the receiving UE, to create the supervisory signal labels $\mathbf{y}$ for our machine learning algorithm.  The downlink BLER for multiple streams for a given UE $q$ is computed as:
\begin{equation}
\text{DL BLER} \triangleq 1- \prod_{j=1}^{n_s} (1 - \text{BLER}_j)
\label{eq:bler}
\end{equation}
where $\text{BLER}_j$ is the observed BLER from stream $j$ for an arbitrary UE $q$.  Thus, $y_i$ is assigned $1$ for a fulfillment of the \textit{hybrid automatic repeat request} (H-ARQ) target $\beta \ge 0$ for the UE $q$ and $y_q$ is assigned $0$ if the BLER exceeded the H-ARQ target for the same UE. The choice of BLER is justified due to its direct relationship with the modulation and code scheme chosen for a given data transmission.  

The choice of these features and supervisory signal enable us to formulate our problem as
\begin{equation}
\begin{aligned}
\underset{n_s}{\textrm{maximize:}} & \qquad \sum_{j=1}^{n_s} \sum_{q=1}^Q C^q_j([\mathbf{X}]_q) \\
\textrm{subject to:} & \qquad n_s \in \{1,\ldots,n_s^\text{max}\}, \\
& \qquad\text{BLER}_q \le \beta, \qquad q = \{1, 2, \ldots, Q\}.
\end{aligned}
\end{equation}
where $C^q_j(\cdot)$ is an unknown function that takes the learning features $\mathbf{X}$ and converts them to a throughput for the user $q$ per radio stream $j$.  This problem is nonconvex due to the nonconvexity of the first constraint.  We therefore resort to machine learning to solve this problem.

The reason why we choose CQI and CSI-RSRP for $\mathbf{X}$ is because they are two physical channel measurement quantities that are not directly correlated:  CSI-RSRP is the received power of the narrowband NR reference symbols while CQI is an indication of the received wideband SINR \cite{3gpp38215}.  If the quantities were correlated or close to correlated, we would have seen an inflation in the training error variance making machine learning inapplicable.  These features are periodically reported to the base station by all the UEs connected.  

\begin {table}[!t]
\setlength\doublerulesep{0.5pt}
\caption{Proposed Machine learning features $\mathbf{X}$ for CoMP}
\vspace*{-0.1in}
\label{table:mlfeatures}
\centering
\begin{tabular}{c|lll} 
\hhline{====}
 & Parameter & Type & Description \\
 \hline
$\mathbf{x_1}$ & CQI & Integer & Linearly transformed CSI-SINR \\
$\mathbf{x_2}$ & CSI-RSRP & Float & CSI-RSRP measurement value \\
\hhline{====}
\end{tabular}
\end{table}

The gathered data $\mathbf{X}$ and $\mathbf{y}$ are periodically split to a training and a test set as part of the proposed machine learning based algorithm.  We then train the model and tune the hyperparameters in Table~\ref{table:hyperparams} using grid search and $K$-fold cross-validation. The SVM classifier used in our algorithm implementation is formulated as an optimization problem
\begin{equation}
\label{eq:svm}
\begin{aligned}
\underset{\bm\lambda}{\textrm{maximize:}}& \,\,\sum_i \lambda_i -  \frac{1}{2} \sum_{n=1}^N\sum_{m =1}^N \lambda_n\lambda_m y_n y_m K(\mathbf{x}_1, \mathbf{x}_2) \\
\textrm{subject to:} & \,\, \sum_{n=1}^N \lambda_ny_n = 0, \\
& \,\, 0 \le \lambda_n \le C, \qquad n = 1, \ldots, N  
\end{aligned}
\end{equation}
where $\lambda_n$ and $\lambda_m$ are elements in $\bm{\lambda}\in \mathbb{R}^N$, which is the Lagrangian multiplier vector resulting from solving the problem using optimization \cite{Cortes95support-vectornetworks}, $C$ is a hyperparameter to control overfitting, also known as the Box constraint. $\mathbf{x}_1$ and $\mathbf{x}_2$ are the first and second feature vectors as defined in Table~\ref{table:mlfeatures}.  Further, $y_n$ is the $n$-th element in the supervisory label  vector $\mathbf{y}\in \{0,1\}^N$. $K(\cdot,\cdot)$ is the SVM kernel and is defined as
\begin{equation}
K(\mathbf{x},\mathbf{x'}) \triangleq \phi(\mathbf{x})^\top\phi(\mathbf{x'})  
\end{equation}
where $\phi(\cdot)$ is a function that maps $\mathbf{x}$ to a higher dimension and $(\cdot)^\top$ is the transpose operation.

What is left is the value size of the collected data $N$, which is defined as
\begin{equation}
N\triangleq \lfloor n_sgQT_\text{sim}/T_\text{CoMP} \rfloor
\label{eq:size}
\end{equation}
where $g$ is the reporting periodicity as defined in \cite{3gpp38211}.  $T_\text{sim}$ and $T_\text{CoMP}$ are the simulation time and CoMP features collection period respectively.

The SVM classifier aims to minimize the \textit{hinge loss} objective $L$, which is a convex loss term as follows
\begin{equation}
\min\colon\qquad L(\mathbf{y},\mathbf{\hat y}) \triangleq \sum_i \max\,(0, 1-y_i\hat{y}_i)
\end{equation}
where $\mathbf{\hat y}\in\{0,1\}^N$ is the predicted supervisory label for CoMP trigger as learned by the classifier.

Since we train the SVM classifier with the training data, we use the test data misclassification error to measure the anticipated performance of the SVM classifier:
\begin{equation}
    \mathrm{Err}\triangleq \frac{1}{N_\text{test}} \sum_{i = 1}^{N_\text{test}} \mathbbm{1}_{(y_i \neq \hat{y}_i)}
\label{eq:misclass}
\end{equation}
where $N_\text{test} \triangleq \lfloor (1 - r_\text{train})N\rfloor$ is the test data size. High misclassification error can be attributed to over-fitting or changed radio conditions.

The problem can be solved with reinforcement learning off-policy solutions such as Q-learning \cite{Sutton}.  However, the problem with using Q-learning is in finding proper initialization of the Q-learning table to avoid exponential runtime \cite{Koenig1993}.  We computed the run time complexity for Q-learning to be $\mathcal{O}(Q^2n_s^\textrm{max})$ \cite{Koenig1993}.

\section{Algorithms}\label{sec:algorithm}

\subsection{Baseline DL CoMP Algorithm}\label{sec:baseline}
Industry recommendations \cite{3gpp36819} suggest physical layer measurements to be used in the formation of the DL CoMP cooperating set.  This is the baseline algorithm. The decision to enable or disable CoMP in the cooperating set for users is based on an absolute minimum threshold of the DL SINR.

\subsection{Improved DL CoMP Algorithm}\label{sec:improved}
\begin{algorithm}[!t]
\small
 \DontPrintSemicolon
 \KwIn{Error threshold $\varepsilon$, prior measurements collection period $T_\mathrm{CoMP}$, current triggering DL SINR, $Q$ UEs reported CQI and CSI-RSRP.  Table~\ref{table:simparameters} has example values.}
\KwOut{Triggering decision for DL CoMP for all $Q$ UEs in $T_\mathrm{sim}$ TTIs.}
\SetKwBlock{Begin}{procedure}{end procedure}

  \For {$T := 1$ \KwTo $T_\mathrm{sim}$} {
     Acquire data $\mathbf{x_1}, \mathbf{x_2}$ from $Q$ UE reports during time $t = T,\ldots,(T + T_\text{CoMP} - 1)$ per Section~\ref{sec:system_model}, which are the learning features $\mathbf{X}$ in Table~\ref{table:mlfeatures}. \;
     Compute the classification label $\mathbf{y}$. \;
    \If {$T\!\!\mod T_{\mathrm{CoMP}} = 0$} {
         Split the data $[\mathbf{X}\,\vert\, \mathbf{y}]$ to training and test data. \;
         Train the SVM model using the training data and use grid search on $K$-fold cross-validation to tune the hyperparameters (in Table~\ref{table:hyperparams}) and compute $\mathbf{\hat{y}}$. \;
         Compute misclassification error $\mathrm{Err}$ as in \eqref{eq:misclass}. \;
          
        \eIf {$\mathrm{Err} \le \varepsilon$}  { 
          
            Decision is to override setting and enable DL CoMP in next TTI if $\textrm{median}(\mathbf{\hat{y}}) = 1$ else disable DL CoMP in next TTI. \;
        }  
        {  
            
              Fallback to operator-entered DL CoMP SINR trigger (baseline algorithm).\;
         }
        Invalidate the SVM model.
    }
}
\caption{\small Improved DL CoMP in heterogeneous networks}\label{alg:algocomp}
\end{algorithm}

The proposed algorithm to trigger CoMP in the cooperating set is shown in Algorithm~\ref{alg:algocomp}.  The error threshold $\varepsilon$ controls the misclassification due to training outside the channel coherence time or sub-optimal fitting. The asymptotic time and space complexity of SVM training is in $\mathcal{O}(M^3)$ and $\mathcal{O}(M^2)$ in the worst case, respectively \cite{Bordes}, where $M$ is the size of the training data ($M \triangleq  \lfloor r_\text{train} N \rfloor$ using \eqref{eq:size}).


\section{Performance Measures} \label{sec:perf}
We use the cumulative distribution of the average UE downlink throughput as follows: peak (95\%), average, and edge (5\%) \cite{link}.  We also use the average BLER as computed in \eqref{eq:bler} and the average number of streams $\bar n_s$, which is equal to the average of the number of used streams $j$ for all the UEs $q$ in the cluster.

\section{Simulation Results} \label{sec:simulation}
\begin{table*}[!t]
\setlength\doublerulesep{0.5pt}
\caption{Radio environment parameters}
\vspace*{-.1in}
\label{table:parameters}
\centering
\begin{threeparttable}
\begin{tabular}{ lr lr}
\hhline{====}
Parameter & Value  & Parameter & Value\\
 \hline
Bandwidth $B$ & 10 MHz & Downlink center frequency & 2100 MHz \\
Channel model type\tnote{\textdagger}& EPA5 & LTE cyclic prefix & Normal \\
Scheduling algorithm & Proportional Fair &  Propagation model & COST231 \\
Propagation environment & Urban & Shadow fading margin standard deviation & 8 dB \\
Macro BS antenna model & Kathrein 742212 &  Maximum number of streams $n_s^\text{max}$  & 2 \\
Pico BS power\tnote{\textasteriskcentered} & 37 dBm & Pico BS antenna height & 10 m\\
Pico BS antenna model & Omnidirectional & Macro BS geometry & Hexagonal \\
Macro BS power & 46 dBm & Macro BS antenna height & 25 m\\
Macro BS antenna electrical tilt & 4$^\circ$ & Inter-site distance & 100 m \\  
UE antenna gain\tnote{*} & -1 dBi & UE height & 1.5 m \\

\hhline{====}
\end{tabular}
\begin{tablenotes}\footnotesize
\item[\textdagger] i.e., the power delay profile.  The UEs are moving at an average speed of 5 km/h.
\item[\textasteriskcentered] BS is short for \textit{base station} and UE is short for \textit{user equipment}.
\end{tablenotes}
\end{threeparttable}
\end{table*}

\begin{table}[!t]
\centering
\setlength\doublerulesep{0.5pt}
\vspace*{-.2in}
\caption{SVM classifier hyperparameters}
\vspace*{-.1in}
\label{table:hyperparams}
\begin{threeparttable}
\begin{tabular}{ll}
\hhline{==}
Hyperparameter & Search range \\
\hline
$K$-fold cross-validation $K$ &  5 \\ 
Training data ratio $r_\text{train}$ & 0.7 \\
Kernel scale & all ranges \\
Box constraint $C$ & all ranges \\
Kernel $K(\cdot, \cdot) $ & \{\texttt{gaussian, linear,} \\
& \texttt{polynomial}\tnote{\textasteriskcentered} \} \\
Normalization & \{\texttt{true, false}\} \\
\hhline{==}
\end{tabular}
\begin{tablenotes}\footnotesize
\item[\textasteriskcentered]{\small Orders 2, 3, and 4.}
\end{tablenotes}
\end{threeparttable}
\end{table}


We use MATLAB and \cite{VLS-2016} to implement our algorithm.  Only the entry point and machine learning codes are available \cite{mycode}.  The simulation parameters are summarized in Table~\ref{table:simparameters}. 

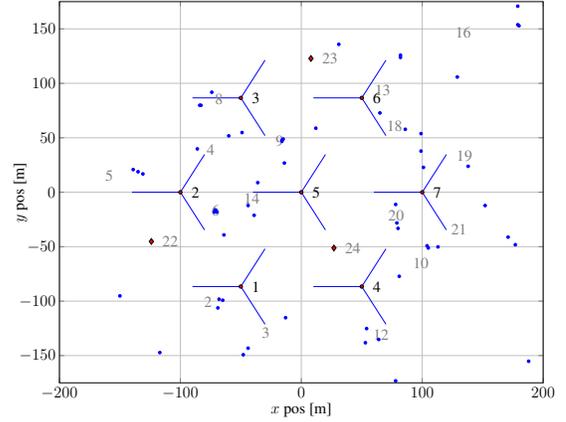
\begin{figure}[!h]
\centering
\scalebox{0.35}{
%
%
\begin{tikzpicture}[font=\Large,scale=1.2]

\begin{axis}[%
width=6.028in,
height=4.754in,
at={(1.011in,0.642in)},
scale only axis,
xmin=-200,
xmax=200,
xtick={-200,-100,...,100,200},
xlabel={$x$ pos [m]},
xmajorgrids,
ymin=-175,
ymax=175,
ylabel={$y$ pos [m]},
ymajorgrids,
axis background/.style={fill=white},
title style={font=\bfseries},
]
\addplot[only marks,mark=*,mark options={},mark size=0.5000pt,draw=white!70!blue,fill=white!70!blue] plot table[row sep=crcr]{%
\\
};
\addplot[only marks,mark=*,mark options={},mark size=0.5000pt,draw=white!70!red,fill=white!70!blue] plot table[row sep=crcr]{%
\\
};
\addplot[only marks,mark=*,mark options={},mark size=1.2000pt,draw=blue,fill=blue] plot table[row sep=crcr]{%
-69	-106.205080756888\\
-65	-99.2050807568877\\
-68	-98.2050807568877\\
-13	-115.205080756888\\
-48	-149.205080756888\\
-44	-143.205080756888\\
-60	51.7949192431123\\
-86	39.7949192431123\\
-49	54.7949192431123\\
-139	20.7949192431123\\
-135	18.7949192431123\\
-131	16.7949192431123\\
-70	-18.2050807568877\\
-71	-16.2050807568877\\
-72	-18.2050807568877\\
-84	79.7949192431123\\
-83	79.7949192431123\\
-74	91.7949192431123\\
-15	48.7949192431123\\
-16	46.7949192431123\\
-16	47.7949192431123\\
104	-49.2050807568877\\
105	-51.2050807568877\\
81	-77.2050807568877\\
54	-125.205080756888\\
53	-138.205080756888\\
64	-135.205080756888\\
82	125.794919243112\\
82	123.794919243112\\
-14	26.7949192431123\\
-44	-12.2050807568877\\
-39	-21.2050807568877\\
-36	8.79491924311228\\
180	152.794919243112\\
179	170.794919243112\\
179	153.794919243112\\
99	37.7949192431123\\
65	72.7949192431123\\
86	57.7949192431123\\
99	53.7949192431123\\
101	22.7949192431123\\
138	23.7949192431123\\
79	-28.2050807568877\\
78	-11.2050807568877\\
80	-33.2050807568877\\
113	-50.2050807568877\\
171	-41.2050807568877\\
177	-48.2050807568877\\
-64	-39.2050807568877\\
-117	-147.205080756888\\
-150	-95.2050807568877\\
129	105.794919243112\\
152	-12.2050807568877\\
31	135.794919243112\\
12	58.7949192431123\\
188	-155.205080756888\\
78	-173.205080756888\\
};
\addplot[only marks,mark=*,mark options={},mark size=0.5000pt,draw=red,fill=blue] plot table[row sep=crcr]{%
\\
};
\addplot [color=blue,solid,forget plot]
  table[row sep=crcr]{%
-50	-86.6025403784439\\
-30	-51.9615242270663\\
};
\addplot [color=blue,solid,forget plot]
  table[row sep=crcr]{%
-50	-86.6025403784439\\
-90	-86.6025403784439\\
};
\addplot [color=blue,solid,forget plot]
  table[row sep=crcr]{%
-50	-86.6025403784439\\
-30	-121.243556529821\\
};
\addplot [color=blue,solid,forget plot]
  table[row sep=crcr]{%
-100	0\\
-80	34.6410161513775\\
};
\addplot [color=blue,solid,forget plot]
  table[row sep=crcr]{%
-100	0\\
-140	0\\
};
\addplot [color=blue,solid,forget plot]
  table[row sep=crcr]{%
-100	0\\
-80	-34.6410161513775\\
};
\addplot [color=blue,solid,forget plot]
  table[row sep=crcr]{%
-50	86.6025403784439\\
-30	121.243556529821\\
};
\addplot [color=blue,solid,forget plot]
  table[row sep=crcr]{%
-50	86.6025403784439\\
-90	86.6025403784439\\
};
\addplot [color=blue,solid,forget plot]
  table[row sep=crcr]{%
-50	86.6025403784439\\
-30	51.9615242270663\\
};
\addplot [color=blue,solid,forget plot]
  table[row sep=crcr]{%
50	-86.6025403784439\\
70	-51.9615242270663\\
};
\addplot [color=blue,solid,forget plot]
  table[row sep=crcr]{%
50	-86.6025403784439\\
10	-86.6025403784439\\
};
\addplot [color=blue,solid,forget plot]
  table[row sep=crcr]{%
50	-86.6025403784439\\
70	-121.243556529821\\
};
\addplot [color=blue,solid,forget plot]
  table[row sep=crcr]{%
0	0\\
20	34.6410161513775\\
};
\addplot [color=blue,solid,forget plot]
  table[row sep=crcr]{%
0	0\\
-40	0\\
};
\addplot [color=blue,solid,forget plot]
  table[row sep=crcr]{%
0	0\\
20	-34.6410161513775\\
};
\addplot [color=blue,solid,forget plot]
  table[row sep=crcr]{%
50	86.6025403784439\\
70	121.243556529821\\
};
\addplot [color=blue,solid,forget plot]
  table[row sep=crcr]{%
50	86.6025403784439\\
10	86.6025403784439\\
};
\addplot [color=blue,solid,forget plot]
  table[row sep=crcr]{%
50	86.6025403784439\\
70	51.9615242270663\\
};
\addplot [color=blue,solid,forget plot]
  table[row sep=crcr]{%
100	0\\
120	34.6410161513775\\
};
\addplot [color=blue,solid,forget plot]
  table[row sep=crcr]{%
100	0\\
60	0\\
};
\addplot [color=blue,solid,forget plot]
  table[row sep=crcr]{%
100	0\\
120	-34.6410161513775\\
};
\addplot[only marks,mark=*,mark options={},mark size=1.5000pt,draw=black,fill=red] plot table[row sep=crcr]{%
-50	-86.6025403784439\\
};
\node[right, align=left, text=black]
at (axis cs:-44,-86.603) {1};
\addplot[only marks,mark=*,mark options={},mark size=1.5000pt,draw=black,fill=red] plot table[row sep=crcr]{%
-100	0\\
};
\node[right, align=left, text=black]
at (axis cs:-94,0) {2};
\addplot[only marks,mark=*,mark options={},mark size=1.5000pt,draw=black,fill=red] plot table[row sep=crcr]{%
-50	86.6025403784439\\
};
\node[right, align=left, text=black]
at (axis cs:-44,86.603) {3};
\addplot[only marks,mark=*,mark options={},mark size=1.5000pt,draw=black,fill=red] plot table[row sep=crcr]{%
50	-86.6025403784439\\
};
\node[right, align=left, text=black]
at (axis cs:56,-86.603) {4};
\addplot[only marks,mark=*,mark options={},mark size=1.5000pt,draw=black,fill=red] plot table[row sep=crcr]{%
0	0\\
};
\node[right, align=left, text=black]
at (axis cs:6,0) {5};
\addplot[only marks,mark=*,mark options={},mark size=1.5000pt,draw=black,fill=red] plot table[row sep=crcr]{%
50	86.6025403784439\\
};
\node[right, align=left, text=black]
at (axis cs:56,86.603) {6};
\addplot[only marks,mark=*,mark options={},mark size=1.5000pt,draw=black,fill=red] plot table[row sep=crcr]{%
100	0\\
};
\node[right, align=left, text=black]
at (axis cs:106,0) {7};
\addplot[only marks,mark=diamond*,mark options={},mark size=2.5883pt,draw=black,fill=red] plot table[row sep=crcr]{%
-124	-45.2050807568877\\
};
\addplot[only marks,mark=diamond*,mark options={},mark size=2.5883pt,draw=black,fill=red] plot table[row sep=crcr]{%
8	122.794919243112\\
};
\addplot[only marks,mark=diamond*,mark options={},mark size=2.5883pt,draw=black,fill=red] plot table[row sep=crcr]{%
27	-51.2050807568877\\
};
\node[align=center, text=gray]
at (axis cs:-77.128,-100.083) {2};
\node[align=center, text=gray]
at (axis cs:-29.315,-128.561) {3};
\node[align=center, text=gray]
at (axis cs:-74.962,39.582) {4};
\node[align=center, text=gray]
at (axis cs:-158.836,15.707) {5};
\node[align=center, text=gray]
at (axis cs:-70.741,-16.39) {6};
\node[align=center, text=gray]
at (axis cs:-67.987,85.336) {8};
\node[align=center, text=gray]
at (axis cs:-17.912,47.462) {9};
\node[align=center, text=gray]
at (axis cs:98.83,-65.444) {10};
\node[align=center, text=gray]
at (axis cs:66.039,-130.666) {12};
\node[align=center, text=gray]
at (axis cs:67.388,93.693) {13};
\node[align=center, text=gray]
at (axis cs:-40.862,-5.943) {14};
\node[align=center, text=gray]
at (axis cs:134.051,146.383) {16};
\node[align=center, text=gray]
at (axis cs:77.252,61.409) {18};
\node[align=center, text=gray]
at (axis cs:134.933,33.448) {19};
\node[align=center, text=gray]
at (axis cs:78.42,-21.17) {20};
\node[align=center, text=gray]
at (axis cs:130.473,-33.84) {21};
\node[right, align=left, text=gray]
at (axis cs:-118,-45.205) {22};
\node[right, align=left, text=gray]
at (axis cs:14,122.795) {23};
\node[right, align=left, text=gray]
at (axis cs:33,-51.205) {24};
\end{axis}
\end{tikzpicture}%
}
\vspace*{-1em}
\caption{Simulated NR network.  The user equipment (UEs) are scattered as blue dots.  The red diamonds are pico base stations (single cell) and the red dots are macro base stations with three cells each (all are numbered).} 
\label{fig:simulated_net_6}
\end{figure}

We run the simulation over a CoMP cooperating set comprised of a single tier of macro BSs with pico BSs scattered in the vicinity.  All macro BSs have three sectors as in Fig.~\ref{fig:simulated_net_6}.
To measure and compare performance of both algorithms, we report the user throughput, which is derived from a respective cumulative distribution function, and the observed BLER based on the average number of streams reported $\bar n_s$.

The standards specify a reporting periodicity $g$ values per TTI \cite{3gpp38211}.   Using default values, and \eqref{eq:size} with the simulation values shown in Table~\ref{table:simparameters}, one collection period has a total of $21240$ samples (collected every $T_\text{CoMP}$ TTIs).

\begin{table}[!t]
\setlength\doublerulesep{0.5pt}
\caption{Simulation parameters}
\vspace*{-.1in}
\label{table:simparameters}
\centering
\begin{threeparttable}
\begin{tabular}{ lr } 
\hhline{==}
Parameter & Value \\
 \hline
Baseline DL CoMP SINR trigger  & 3 dB \\
Number of cooperating cells per cluster & 32 \\
Total number of UEs $Q$ in the cluster & 60 \\
Number of pico BSs per cluster & 11 \\
H-ARQ target $\beta$ & 0.1 \\
NR frame duration & 10 TTIs \\
Features collection period $T_\text{CoMP}$ & 3 TTIs \\
Simulation time $T_\text{sim}$ & 60 TTIs \\
Error threshold $\varepsilon$ & 12\% \\
 \hhline{==}
\end{tabular}
\end{threeparttable}
\end{table}

\begin{figure}[!t]
\centering
\subfloat[Baseline]{\scalebox{0.24}{
%
%
\begin{tikzpicture}
\tikzstyle{every node}=[font=\Large,scale=2]
\begin{axis}[%
width=6.028in,
height=4.754in,
at={(1.011in,0.642in)},
scale only axis,
xmin=0,
xmax=60,
xlabel={TTI},
xmajorgrids,
ymin=-0.5,
ymax=1.5,
ytick={0,1},
ylabel={CoMP Decision},
ymajorgrids,
axis background/.style={fill=white},
]
\addplot [color=black,solid,forget plot]
  table[row sep=crcr]{%
0	1\\
1	1\\
2	1\\
3	1\\
4	1\\
5	1\\
6	1\\
7	1\\
8	0\\
9	0\\
10	0\\
11	0\\
12	0\\
13	0\\
14	0\\
15	0\\
16	0\\
17	0\\
18	0\\
19	0\\
20	0\\
21	0\\
22	0\\
23	0\\
24	0\\
25	0\\
26	0\\
27	0\\
28	0\\
29	0\\
30	0\\
31	0\\
32	0\\
33	0\\
34	0\\
35	0\\
36	0\\
37	0\\
38	0\\
39	0\\
40	0\\
41	0\\
42	0\\
43	0\\
44	0\\
45	0\\
46	0\\
47	0\\
48	0\\
49	0\\
50	0\\
51	0\\
52	0\\
53	0\\
54	0\\
55	0\\
56	0\\
57	0\\
58	0\\
59	0\\
60	0\\
};
\end{axis}
\end{tikzpicture}%
}}
\hfil
\subfloat[Proposed]{\scalebox{0.24}{
%
%
\begin{tikzpicture}
\tikzstyle{every node}=[font=\Large,scale=2]
\begin{axis}[%
width=6.028in,
height=4.754in,
at={(1.011in,0.642in)},
scale only axis,
xmin=0,
xmax=60,
xlabel={TTI},
xmajorgrids,
ymin=-0.5,
ymax=1.5,
ytick={0,1},
ymajorgrids,
axis background/.style={fill=white},
]
\addplot [color=black,solid]
  table[row sep=crcr]{%
0	1\\
1	1\\
2	1\\
3	1\\
4	1\\
5	1\\
6	1\\
7	1\\
8	0\\
9	0\\
10	0\\
11	0\\
12	0\\
13	1\\
14	1\\
15	1\\
16	1\\
17	1\\
18	1\\
19	1\\
20	1\\
21	1\\
22	1\\
23	1\\
24	1\\
25	1\\
26	1\\
27	1\\
28	1\\
29	0\\
30	0\\
31	0\\
32	1\\
33	1\\
34	1\\
35	1\\
36	1\\
37	1\\
38	1\\
39	1\\
40	1\\
41	1\\
42	0\\
43	0\\
44	0\\
45	0\\
46	0\\
47	1\\
48	1\\
49	1\\
50	1\\
51	1\\
52	1\\
53	1\\
54	1\\
55	1\\
56	0\\
57	0\\
58	0\\
59	0\\
60	0\\
};

\end{axis}
\end{tikzpicture}%
}}
\caption{Downlink coordinated multipoint (DL CoMP) being enabled \mbox{(state = 1)} and disabled (state = 0) for both baseline (left) and the proposed algorithm (right) over the same transmit time intervals (TTI).}
\label{fig:sinrcomp6}
\end{figure}
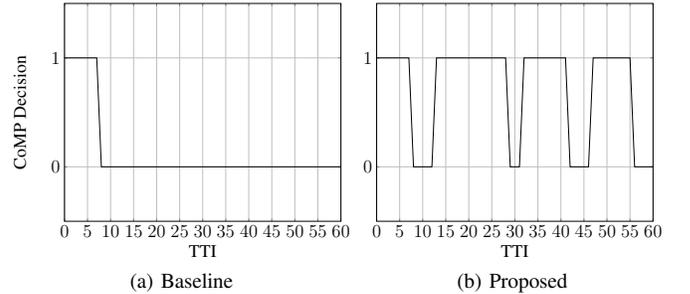

\begin {table}[!t]
\setlength\doublerulesep{0.5pt}
\caption{Throughput Measures for Downlink Coordinated Multipoint}
\vspace*{-.1in}
\label{table:kpis6}
\begin{adjustwidth}{-.1in}{0cm}
\centering
\begin{tabular}{ p{0.07\textwidth}ccc|ccc } 
 & \multicolumn{6}{c}{User Equipment Throughput [Mbps]} \\
 \cline{2-7}
  & \multicolumn{3}{c}{Baseline\tnote{\textdagger}} & \multicolumn{3}{c}{Dynamic} \\
\cline{2-7}
Cluster & Peak & Average & Edge  & Peak & Average & Edge  \\
\hline
Macro & 1.73 & 0.77 & 0.01 & \textbf{1.83} & 0.77 & 0.01 \\
Pico & 2.59 & 1.63 & 0.12 & \textbf{3.36} & \textbf{1.88} & \textbf{0.22} \\
Overall & 2.13 & 0.91 & 0.02 & \textbf{2.67} & \textbf{0.94} & 0.02 \\
\hhline{=======}
\end{tabular}
\end{adjustwidth}
\end{table}

\begin {table}[!t]
\setlength\doublerulesep{0.5pt}
\caption{Link-level Measures for Downlink Coordinated Multipoint}
\vspace*{-.1in}
\label{table:linklevel6}
\centering
\begin{tabular}{ p{0.12\textwidth}cccc} 
  & \multicolumn{4}{c}{Average} \\
\cline{2-5}
Scenario & DL BLER & $\bar n_s$ & CQI & CSI-RSRP [dBm] \\
 \hline
Baseline\tnote{\textdagger}  & 15.89\% & 1.58 & 4 & -66.74 \\
Dynamic CoMP & 15.97\% & 1.59 & 4 & -66.74 \\
\hhline{=====}
\end{tabular}
\end{table}

In Fig. \ref{fig:sinrcomp6}, the baseline algorithm made decisions to enable or disable CoMP in the cooperating set of users where the improved dynamic algorithm made the opposite decision. Tables~\ref{table:kpis6}~and~\ref{table:linklevel6} outline the performance measures and show that the proposed CoMP algorithm shows improved UE throughput with no change in the CSI-RSRP or CQI.  The reason for the throughput improvement is the improved CoMP triggering conditions, which are used to dynamically reconfigure the number of transmit streams (from the BS side) for the UEs in the CoMP cooperating set with no change in the total transmit power.  The BLER is expected to increase with the increase in the number of transmit streams $n_s$.  However, the overall MIMO gain due to CoMP triggering of a second stream exceeds the loss in performance due to the increased BLER. 

\section{Conclusions}\label{sec:conclusion}
In this paper, we used online machine learning and physical layer measurements to train an SVM classifier.  The measurements were collected and used within the channel coherence time.  The was model invalidated after the coherence time passed.  This approach improved the CoMP joint processing distributed MIMO performance by transmitting another spatially uncorrelated stream.  We did so without compromising the reported CQI or received power.   We only used two learning features for SVM and showed that they were sufficient.  We used the fulfillment of the H-ARQ target as our supervisory signal. We chose a realistic heterogeneous environment with no channel reciprocity. Our simulated results showed improvement in the user throughput distribution compared to the baseline CoMP algorithm.

\bibliography{references}  
\bibliographystyle{IEEEtran}

%

%
%




\end{document}